\documentclass{article} 
\usepackage{iclr2017_conference,times}
\usepackage{hyperref}
\usepackage{url}
\usepackage{graphicx}
\usepackage{amssymb}
\usepackage{amsmath}
\usepackage{subfig}
\usepackage{bm}
\usepackage{wrapfig}

\iclrfinalcopy
\title{Attend, Adapt and Transfer:\\Attentive Deep Architecture for Adaptive Transfer from multiple sources in the same domain}

\author{Janarthanan Rajendran \thanks{Authors contributed equally}\\
University of Michigan\\
\texttt{rjana@umich.edu} \\
\And
Aravind ~S. Lakshminarayanan \footnotemark[1]\\
Indian Institute of Technology Madras\\
\texttt{aravindsrinivas@gmail.com} \\
\And
Mitesh M. Khapra\\
Indian Institute of Technology Madras\\
\texttt{miteshk@cse.iitm.ac.in}\\
\And
Prasanna P\\
McGill University\\
\texttt{prasanna.p@cs.mcgill.ca}\\
\And
Balaraman Ravindran\\
Indian Institute of Technology Madras\\
\texttt{ravi@cse.iitm.ac.in}
}

\begin{document}

\maketitle

\begin{abstract}
Transferring \textit{knowledge} from prior source tasks in solving a new target task can be useful in several learning applications.
The application of transfer poses two serious challenges which have not been adequately addressed. First, the agent should be able to avoid negative transfer, which happens when the transfer hampers or slows down the learning instead of helping it. Second, the agent should be able to selectively transfer, which is the ability to select and transfer from different and multiple source tasks for different parts of the state space of the target task. We propose A2T (Attend, Adapt and Transfer), an attentive deep architecture which adapts and transfers from these source tasks. Our model is generic enough to effect transfer of either policies or value functions. Empirical evaluations on different learning algorithms show that A2T is an effective architecture for transfer by being able to avoid negative transfer while transferring selectively from multiple source tasks in the same domain.

\end{abstract}

\section{Introduction}
\label{intro}

One of the goals of Artificial Intelligence (AI) is to build autonomous agents that can learn and adapt to new environments. Reinforcement Learning (RL) is a key technique for achieving such adaptability. The goal of RL algorithms is to learn an optimal policy for choosing actions that maximize some notion of long term performance. Transferring knowledge gained from tasks solved earlier to solve a new target task can help, either in terms of speeding up the learning process or in terms of achieving a better solution, among other performance measures. When applied to RL, transfer could be accomplished in many ways (see \cite{pstonesurvey,taylorsurvey} for a very good survey of the field). One could use the value function from the source task as an initial estimate in the target task to cut down exploration [\cite{sorgtransfer}]. Alternatively one could use policies from the source task(s) in the target task. This can take one of two forms - (i) the derived policies can be used as initial exploratory trajectories [\cite{lfd,lfd2}] in the target task and (ii) the derived policy could be used to define {\it macro-actions} which may then be used by the agent in solving the target task [\cite{mannoroptions,brunskill}].

While transfer in RL has been much explored, there are two crucial issues that have not been adequately addressed in the literature. The first is {\it negative transfer}, which occurs when the transfer results in a performance that is worse when compared to learning from scratch in the target task. This severely limits the applicability of many transfer techniques only to cases for which some measure of relatedness between source and target tasks can be guaranteed beforehand. This brings us to the second problem with transfer, which is the issue of identifying an appropriate source task from which to transfer. In some scenarios, different source tasks might be relevant and useful for different parts of the state space of the target task. As a real world analogy, consider multiple players (experts) who are good at different aspects of a game (say, tennis). For example, Player 1 is good at playing backhand shots while Player 2 is good at playing forehand shots. Consider the case of a new player (agent) who wants to learn tennis by selectively learning from these two experts. We handle such a situation in our architecture by allowing the agent to learn how to pick and use solutions from multiple and different source tasks while solving a target task, selectively applicable for different parts of the state space. We call this {\it selective transfer}. Our agent can transfer knowledge from Player 1 when required to play backhand shots and Player 2 for playing forehand shots. Further, let us consider consider the situation that both Player 1 and Player 2 are bad at playing drop shots. Apart from the source tasks, we maintain a base network that learns from scratch on the target task. The agent can pick and use the solution of the base network when solving the target task at the parts of the state space where transferring from the source tasks is negative. Such a situation could arise when the source task solutions are irrelevant for solving the target task over a specific portion of the state space, or when the transferring from the source tasks is negative over a specific portion of the state space (for example, transferring the bad drop shot abilities of Players 1 and 2).  This situation also entails the first problem of avoiding negative transfer. Our framework allows an agent to avoid transferring from both Players 1 and 2 while learning to play drop shots, and rather acquire the drop shot skill by learning to use the base network. The architecture is trained such that the base network uses not just the experience obtained through the usage of its solutions in the target task, but the overall experience acquired using the combined knowledge of the source tasks and itself. This enables the base network solutions to get closer to the behavior of the overall architecture (which uses the source task solutions as well). This makes it easier for the base network to assist the architecture to fine tune the useful source task solutions to suit the target task perfectly over time. 

The key contribution in the architecture is a {\it deep {\textbf{attention}} network}, that decides which solutions to attend to, for a given input state. The network learns  solutions as a function of current state thereby aiding the agent in adopting different solutions for different parts of the state space in the target task. 

To this end, we propose A2T: Attend, Adapt and Transfer, an Attentive Deep Architecture for Adaptive Transfer, that avoids negative transfer while performing selective transfer from multiple source tasks in the same domain. In addition to the tennis example, A2T is a fairly generic framework that can be used to selectively transfer different skills available from different experts as appropriate to the situation. For instance, a household robot can appropriately use skills from different experts for different household chores. This would require the skill to transfer manipulation skills across objects, tasks and robotic actuators. With a well developed attention mechanism, the most appropriate and helpful combination of object-skill-controller can be identified for aiding the learning on a related new task.
Further, A2T is generic enough to effect transfer of either action policies or action-value functions, as the case may be. We also adapt different algorithms in reinforcement learning as appropriate for the different settings and empirically demonstrate that the A2T is effective for transfer learning for each setting.


\section{Related Work}
\label{related_work}


As mentioned earlier, transfer learning approaches could deal with transferring policies or value functions. For example, \cite{bikramjit} describe a method for transferring value functions by constructing a {\it Game tree}. Similarly, \cite{sorgtransfer} use the value function from a source task as the initial estimate of the value function in the target task. 

Another method to achieve transfer is to reuse policies derived in the source task(s) in the target task. Probabilistic Policy Reuse as discussed in \cite{policyreuse} maintains a library of policies and selects a policy based on a similarity metric, or a random policy, or a max-policy from the knowledge obtained. This is different from the proposed approach in that the proposed approach can transfer policies at the granularity of individual states which is not possible in policy-reuse rendering it unable to learn customized policy at that granularity.
\cite{lfd,lfd2} evaluated the idea of having the transferred policy from the source tasks as explorative policies instead of having a random exploration policy. This provides better exploration behavior provided the tasks are similar. \cite{talvitie} try to find the promising policy from a set of candidate policies that are generated using different action mapping to a single solved task. In contrast, we make use of one or more source tasks to selectively transfer policies at the granularity of state. 
Apart from policy transfer and value transfer as discussed above, \cite{sridharproto} discuss representation transfer using Proto Value Functions.
 
The idea of negative and selective transfer have been discussed earlier in the literature. For example,  \cite{lazarictransfer} address the issue of negative transfer in transferring samples for a related task in a multi-task setting.  \cite{andytransfer} discuss the idea of exploiting shared common features across related tasks. They learn a {\it shaping function} that can be used in later tasks. 


The two recent works that are very relevant to the proposed architecture are discussed in \cite{actormimic} and \cite{progressivenets}. \cite{actormimic} explore transfer learning in RL across Atari games by trying to learn a multi-task network over the source tasks available and directly fine-tune the learned multi-task network on the target task. However, fine-tuning as a transfer paradigm cannot address the issue of negative transfer which they do observe in many of their experiments. \cite{progressivenets} try to address the negative transfer issue 
by proposing a sequential learning mechanism where the filters of the network being learned for an ongoing task are dependent through lateral connections on the lower level filters of the networks learned already for the previous tasks. 
The idea is to ensure that dependencies that characterize similarity across tasks could be learned through these lateral connections. Even though they do observe better transfer results than direct fine-tuning, they are still not able to {\it avoid} negative transfer in some of their experiments. 


\section{Proposed Architecture}
\label{model}
Let there be $N$ source tasks and let $K_1, K_2, \ldots K_N$ be the solutions of these source tasks $1,\ldots N$ respectively. Let $K_T$ be the solution that we learn in the target task $T$. Source tasks refer to tasks that we have already learnt to perform and target task refers to the task that we are interested in learning now. These solutions could be for example policies or state-action values. Here the source tasks should be in the same domain as the target task, having the same state and action spaces. We propose a setting where $K_T$ is learned as a function of $K_1,\ldots,K_N,K_B$, where $K_{B}$ is the solution of a base network which starts learning from scratch while acting on the target task.
In this work, we use a convex combination of the solutions to obtain $K_T$.
\begin{equation}
K_{T}(s) = w_{N+1,s} K_{B}(s) + \sum_{i=1}^{N} w_{i,s} K_{i}(s)
\label{eq:comb}
\end{equation}
\begin{equation}
\sum_{i=1}^{N+1} w_{i,s} = 1 , w_{i,s} \in [0,1]
\label{eq:weight_cond}
\end{equation}
$w_{i,s}$ is the weight given to the $i$th solution at state $s$. 

\begin{figure}
\centering
\subfloat[]{
        \label{fig:2m2}
        \includegraphics[width=0.5\textwidth]{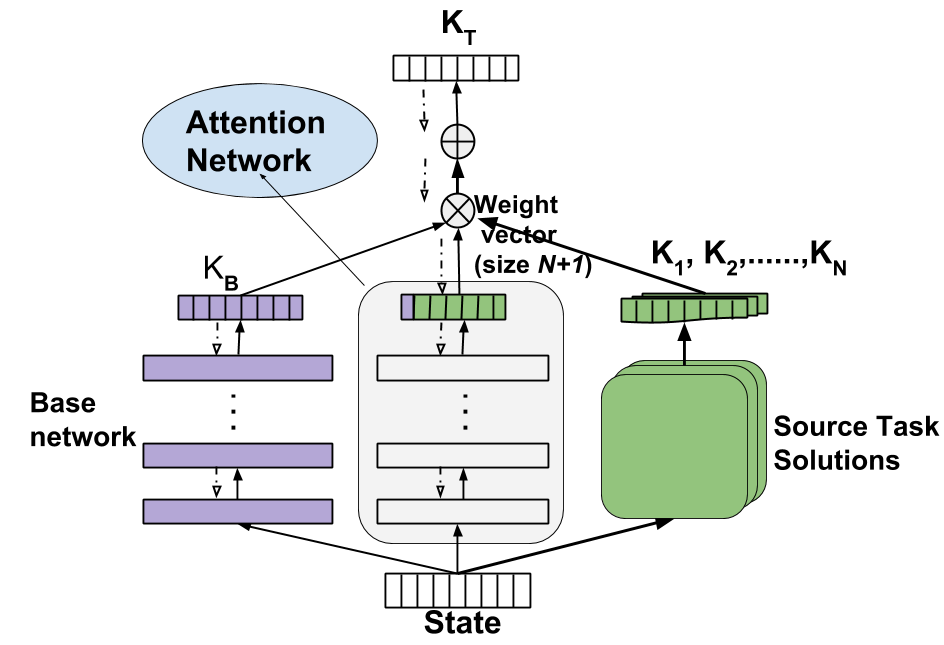}}
\subfloat[]{
        \label{fig:actor_critic}
        \includegraphics[width=0.4\textwidth]{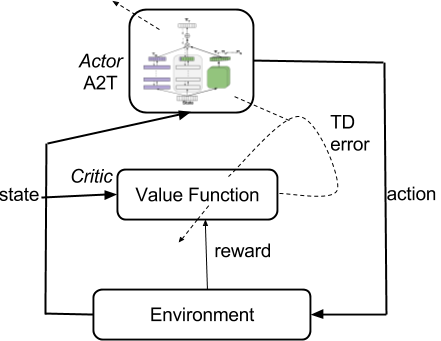}}

\caption{(a) A2T architecture. The doted arrows represent the path of back propagation. (b) Actor-Critic using A2T.}
\end{figure}

The agent uses $K_{T}$ to act in the target task. Figure \ref{fig:2m2} shows the proposed architecture. While the source task solutions $K_1, \ldots, K_N$ remain fixed, the base network solutions are learnt and hence $K_B$ can change over time. There is a central network which learns the weights ($w_{i,s}$, $i \in 1,2,\ldots,N+1$), given the input state $s$. We refer to this network as the {\it attention network}. The $[0,1]$ weights determine the attention each solution gets allowing the agent to selectively accept or reject the different solutions, depending on the input state. We adopt a {\it soft-attention} mechanism whereby more than one weight can be non-zero [\cite{RLattention}] as opposed to a {\it hard-attention} mechanism [\cite{mnih2014recurrent}] where we are forced to have only one non-zero weight. 

\begin{equation}
w_{i,s} = \frac{\exp{(e_{i,s})}}{\sum\limits_{j=1}^{N+1} \exp{(e_{j,s})}}, i \in \{1,2, \ldots,N+1\}
\label{eq:att}
\end{equation}
\begin{equation}
 (e_{1,s}, e_{2,s}, \ldots , e_{N+1,s}) = f(s;\theta_a)
\label{eq:att_net_eq}
\end{equation}

Here, $f(s;\theta_a)$ is a deep neural network (attention network), which could consist of convolution layers and fully connected layers depending on the representation of input. It is parametrised by $\theta_a$ and takes as input a state $s$ and outputs a vector of length $N+1$, which gives the attention scores for the $N+1$ solutions at state $s$. Eq.(\ref{eq:att}) normalises this score to get the weights that follow Eq.(\ref{eq:weight_cond}).

If the $i$th source task solution is useful at state $s$, then $w_{i,s}$ is set to a high value by the attention network. Working at the granularity of states allows the attention network to attend to different source tasks, for different parts of the state space of the target task, thus giving it the ability to perform selective transfer. For parts of the state space in the target task, where the source task solutions cause negative transfer or where the source task solutions are not relevant, the attention network learns to give high weight to the base network solution (which can be learnt and improved), thus avoiding negative transfer.


Depending on the feedback obtained from the environment upon following $K_{T}$, the attention network's parameters $\theta_a$ are updated to improve performance.

As mentioned earlier, the source task solutions, $K_{1}, \ldots, K_{N}$ remain fixed. Updating these source task's parameters would cause a significant amount of unlearning in the source tasks solutions and result in a weaker transfer, which we observed empirically. This also enables the use of source task solutions, as long as we have the outputs alone, irrespective of how and where they come from.

Even though the agent follows $K_{T}$, we update the parameters of the base network that produces $K_{B}$, as if the action taken by the agent was based only on $K_{B}$. 
Due to this special way of updating $K_B$, apart from the experience got through the unique and individual contribution of $K_B$ to $K_T$ in parts of the state space where the source task solutions are not relevant, $K_B$ also uses the valuable experience got by using $K_T$ which uses the solutions of the source tasks as well. 

This also means that, if there is a source task whose solution $K_{j}$ is useful for the target task in some parts of its state space, then $K_{B}$ tries to replicate $K_{j}$ in those parts of the state space.
In practise, the source task solutions though useful, might need to be modified to suit perfectly for the target task. The base network takes care of these modifications required to make the useful source task solutions perfect for the target task. The special way of training the base network assists the architecture in achieving this faster.
Note that the agent could follow/use $K_{j}$ through $K_T$ even when $K_{B}$ does not attain its replication in the corresponding parts of the state space.  
This allows for a good performance of the agent in earlier stages training itself, when a useful source task is available and identified.

Since the attention is soft, our model has the flexibility to combine multiple solutions. 
The use of deep neural networks allow the model to work even for large, complex RL problems. The deep attention network, allows the agent to learn complex selection functions, without worrying about representation issues a priori. To summarise, for a given state, A2T learns to {\it attend} to specific solutions and {\it adapts} this attention over different states, hence attaining useful {\it transfer}. A2T is general and can be used for transfer of solutions such as policy and value.

\subsection{Policy Transfer}
The solutions that we transfer here are the source task policies, taking advantage of which, we learn a policy for the target task. Thus, we have $K_1, \ldots, K_N, K_B, K_T \leftarrow  \pi_1, \ldots \pi_N, \pi_B, \pi_T$. Here $\pi$ represents a stochastic policy, a probability distribution over all the actions. The agent acts in the target task, by sampling actions from the probability distribution $\pi_T$. The target task policy $\pi_T$ is got as described in Eq.(\ref{eq:comb}) and Eq.(\ref{eq:weight_cond}). The attention network that produces the weights for the different solutions, is trained by the feedback got after taking action following $\pi_T$. The base network that produces $\pi_B$ is trained as if the sampled action came from $\pi_B$ (though it originally came from $\pi_T$), the implications of which were discussed in the previous section.
When the attention network's weight for the policy $\pi_{B}$ is high, the mixture policy $\pi_T$ is dominated by $\pi_{B}$, and the base network learning is nearly on-policy. In the other cases, $\pi_{B}$ undergoes off-policy learning. But if we look closely, even in the latter case, since $\pi_B$ moves towards $\pi_T$, it tries to be nearly on-policy all the time. Empirically, we observe that $\pi_{B}$ converges. This architecture for policy transfer can be used alongside any algorithm that has an explicit representation of the policy. Here we describe two instantiations of A2T for policy transfer, one for direct policy search using REINFORCE algorithm and another in the Actor-Critic setup. 

\subsubsection{Policy Transfer in REINFORCE Algorithms using A2T:}
REINFORCE algorithms [\cite{reinforce}] can be used for direct policy search by making weight adjustments in a direction that lies along the gradient of the expected reinforcement. The full architecture is same as the one shown in Fig.\ref{fig:2m2} with $K \leftarrow \pi$. We do direct policy search, and the parameters are updated using REINFORCE. Let the attention network be parametrized by $\theta_a$ and the base network which outputs $\pi_B$ be parametrized by $\theta_b$. The updates are given by:

\begin{equation}
\theta_a \leftarrow \theta_a +  \alpha_{\theta_a}(r - b)\frac{\partial \sum_{t=1}^{M}\log(\pi_T(s_t,a_t))}{\partial \theta_a}
\end{equation}
\begin{equation}
\theta_b \leftarrow \theta_b + \alpha_{\theta_b}(r - b)\frac{\partial \sum_{t=1}^{M}\log(\pi_{B}(s_t,a_t))}{\partial \theta_b}
\end{equation}
where $\alpha_{\theta_a}, \alpha_{\theta_b} $ are non-negative factors, $r$ is the return obtained in the episode, $b$ is some baseline and $M$ is the length of the episode. $a_t$ is the action sampled by the agent at state $s_t$ following $\pi_T$. Note that while $\pi_T(s_t,a_t)$ is used in the update of the attention network, $\pi_B(s_t,a_t)$ is used in the update of the base network. 

\subsubsection{Policy Transfer in Actor-Critic using A2T:}
Actor-Critic methods [\cite{KondaAC}] are Temporal Difference (TD) methods that have two separate components, \textit{viz.}, an {\it actor} and a {\it critic}. The {actor} proposes a policy whereas the critic estimates the value function to critique the actor's policy. The updates to the actor happens through {\it TD-error} which is the one step estimation error that helps in reinforcing an agent's behaviour.

We use A2T for the actor part of the Actor-Critic. The architecture is shown in Fig.\ref{fig:actor_critic}. 
The actor, A2T is aware of all the previous learnt tasks and tries to use those solution policies for its benefit. The critic evaluates the action selection from $\pi_T$ on the basis of the performance on the target task. With the same notations as REINFORCE for $s_t, a_t, \theta_a, \theta_b, \alpha_{\theta_a}, \alpha_{\theta_b}, \pi_B, \pi_T$; let action $a_t$ dictated by $\pi_T$ lead the agent to next state $s_{t+1}$ with a reward $r_{t+1}$ and let $V(s_t)$ represent the value of state $s_t$ and $\gamma$ the discount factor. Then, the update equations for the actor are as below:
\begin{equation}
\delta_t = r_{t+1} + \gamma V(s_{t+1}) - V(s_t) 
\end{equation}
\begin{equation}
\theta_a \leftarrow \theta_a + \alpha_{\theta_a} \delta_t \frac{\frac{\partial \log \pi_T(s_t, a_t)}{\partial \theta_a}}{\left|\frac{\partial \log \pi_T(s_t, a_t)}{\partial \theta_a}\right|} 
\end{equation}
\begin{equation}
\theta_b \leftarrow \theta_b + \alpha_{\theta_b} \delta_t \frac{\frac{\partial \log \pi_{B}(s_t, a_t)}{\partial \theta_b}}{\left|\frac{\partial \log \pi_{B}(s_t, a_t)}{\partial \theta_b}\right|} 
\end{equation}
Here, $\delta_t$ is the TD error. The state-value function $V$ of the critic is learnt using TD learning.




\subsection{Value Transfer}
In this case, the solutions being transferred are the source tasks' action-value functions, which we will call as $Q$ functions. Thus, $K1,\ldots, K_N, K_B, K_T \leftarrow Q_1,\ldots, Q_N, Q_B, Q_T$. Let $A$ represent the discrete action space for the tasks and $Q_i(s) = \{Q(s,a_j) \ \forall \ a_j \in A\}$. The agent acts by using $Q_T$ in the target task, which is got as described in Eq.(\ref{eq:comb}) and Eq.(\ref{eq:weight_cond}). The attention network and the base network of A2T are updated as described in the architecture. 



\subsubsection{Value Transfer in Q learning using A2T:}
The state-action value $Q$ function is used to guide the agent to selecting the optimal action $a$ at a state $s$, where $Q(s,a)$ is a measure of the long-term return obtained by taking action $a$ at state $s$. One way to learn optimal policies for an agent is to estimate the optimal $Q(s,a)$ for the task. Q-learning [\cite{Watkins}] is an off-policy Temporal Difference (TD) learning algorithm that does so. The Q-values are updated iteratively through the Bellman optimality equation [\cite{puterman}] with the rewards obtained from the task as below:
$$Q(s,a) \leftarrow \mathbb{E}[r(s,a,s') + \gamma \textrm{max}_{a'} Q(s',a')]$$ 

In high dimensional state spaces, it is infeasible to update Q-value for all possible state-action pairs. One way to address this issue is by approximating $Q(s,a)$ through a parametrized function approximator $Q(s,a;\theta)$,thereby generalizing over states and actions by operating on higher level features [\cite{rlbook}]. The DQN [\cite{mnih2015human}] approximates the Q-value function with a deep neural network to be able to predict $Q(s,a)$ over all actions $a$, for all states $s$. 

The loss function used for learning a Deep Q Network is as below:
$$L(\theta) = \mathbb{E}_{s,a,r,s'}[\left(y^{DQN}- Q(s,a;\theta)\right)^2],$$
with
$$y^{DQN} = \left(r+ \gamma \textrm{max}_{a'}Q(s',a',\theta^{-})\right)$$
Here, $L$ represents the expected TD error corresponding to current parameter estimate $\theta$. $\theta^{-}$ represents the parameters of a separate {\it target network}, while $\theta$ represents the parameters of the {\it online network}. The usage of a {\it target network} is to improve the stability of the learning updates. The gradient descent step is shown below:
$$\nabla_{\theta} L(\theta) = \mathbb{E}_{s,a,r,s'} [(y^{DQN} - Q(s,a;\theta))\nabla_{\theta} Q(s,a)]$$
To avoid correlated updates from learning on the same transitions that the current network simulates, an experience replay [\cite{Lin}] $D$ (of fixed maximum capacity) is used, where the experiences are pooled in a FIFO fashion. 


We use DQN to learn our experts $Q_i, i \in {1,2 \ldots N}$ on the source tasks. Q-learning is used to ensure $Q_T(s)$ is driven to a good estimate of $Q$ functions for the target task. Taking advantage of the off-policy nature of Q-learning, both $Q_B$ and $Q_T$ can be learned from the experiences gathered by an $\epsilon$-greedy behavioral policy based on $Q_T$. Let the attention network that outputs $w$ be parametrised by $\theta_a$ and the base network outputting $Q_{B}$ be parametrised by $\theta_b$. Let ${\theta_a}^{-}$ and ${\theta_b}^{-}$ represent the parameters of the respective target networks. Note that the usage of {\it target} here is to signify the parameters ($\theta_a^{-}, \theta_b^{-}$) used to calculate the {\it target} value in the Q-learning update and is different from its usage in the context of the {\it target} task. The update equations are:
\begin{equation}
y^{Q_T} = (r+ \gamma \textrm{max}_{a'}Q_T(s',a';{\theta_{a}}^{-}, {\theta_{b}}^{-}))
\end{equation}
\begin{equation}
L^{Q_T}(\theta_{a}, \theta_{b}) = \mathbb{E}_{s,a,r,s'}[(y^{Q_T}- Q_T(s,a;\theta_{a},\theta_{b}))^2]
\label{eq:att_q}
\end{equation}
\begin{equation}
L^{Q_B}(\theta_{b}) = \mathbb{E}_{s,a,r,s'}[(y^{Q_T}- Q_B(s,a;\theta_{b}))^2]
\label{eq:base_q}
\end{equation}
\begin{equation}
\nabla_{\theta_{a}} L^{Q_T} = \mathbb{E} [(y^{Q_T} - Q_T(s,a))\nabla_{\theta_{a}} Q_T(s,a)]
\end{equation}
\begin{equation}
\nabla_{\theta_{b}} L^{Q_B} = \mathbb{E} [(y^{Q_T} - Q_B(s,a))\nabla_{\theta_{b}} Q_R(s,a)]
\end{equation}
$\theta_a$ and $\theta_b$ are updated with the above gradients using RMSProp. Note that the Q-learning updates for both the attention network (Eq.(\ref{eq:att_q})) and the base network (Eq.(\ref{eq:base_q})) use the target value generated by $Q_T$. We use {\it target} networks for both $Q_B$ and $Q_T$ to stabilize the updates and reduce the non-stationarity as in DQN training. The parameters of the {\it target} networks are periodically updated to that of the {\it online} networks. 


\section{Experiments and Discussion}
\begin{figure}
\centering
\subfloat[Chain World]{
        \label{fig:chain_w}
        \includegraphics[width=0.3\textwidth]{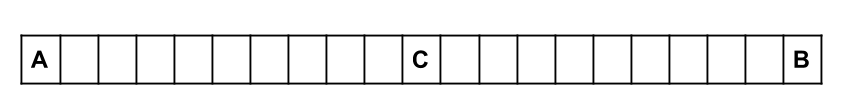}}
\subfloat[Puddle World 1]{
        \label{fig:w1}
        \includegraphics[width=0.31\textwidth]{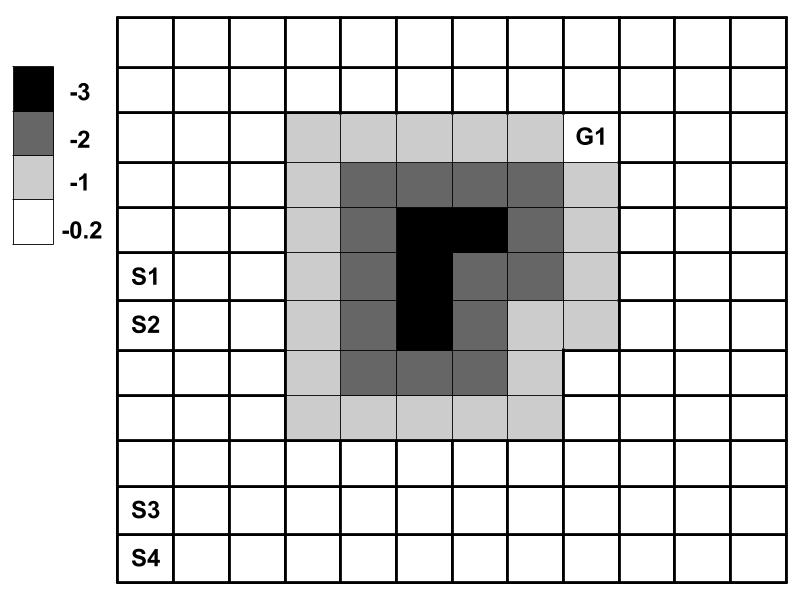} }
\subfloat[Puddle World 2]{
        \label{fig:w2}
        \includegraphics[width=0.31\textwidth]{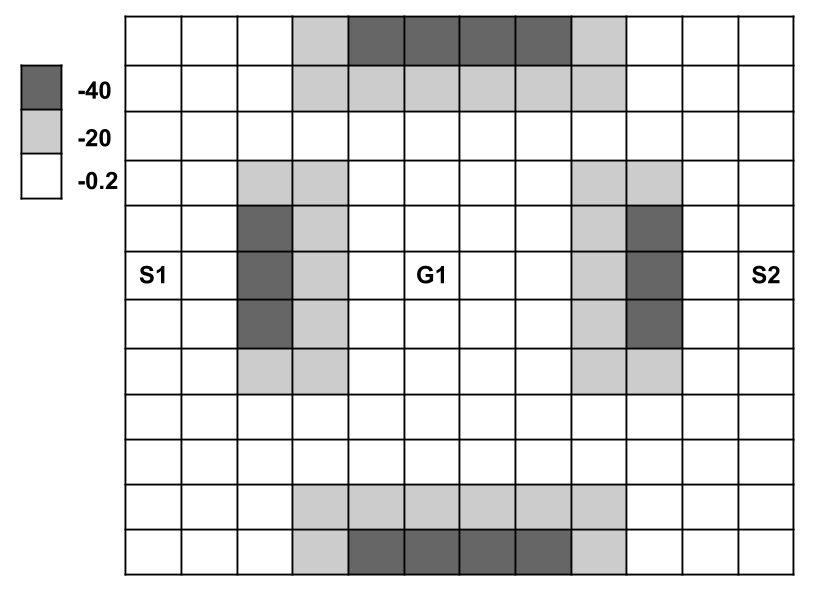} }
\caption{Different worlds for policy transfer experiments}
\label{kohler}
\end{figure}

We evaluate the performance of our architecture A2T on policy transfer using two simulated worlds, \textit{viz.}, chain world and puddle world as described below. The main goal of these experiments is to test the consistency of results with the algorithm motivation.
\textbf{Chain world:} Figure \ref{fig:chain_w} shows the chain world where the goal of the agent is to go from one point in the chain (starting state) to another point (goal state) in the least number of steps. At each state the agent can choose to either move one position to the left or to the right. After reaching the goal state the agent gets a reward that is inversely proportional to the number of steps taken to reach the goal. 

\textbf{Puddle worlds:} Figures \ref{fig:w1} and \ref{fig:w2} show the discrete version of the standard puddle world that is widely used in Reinforcement Learning literature. In this world, the goal of the agent is to go from a specified start position to the goal position, maximising its return. At each state the agent can choose one of these four actions: move one position to the north, south, east or west.With $0.9$ probability the agent moves in the chosen direction and with $0.1$ probability it moves in a random direction irrespective of its choice of action. On reaching the goal state, the agent gets a reward of $+10$. On reaching other parts of the grid the agent gets different penalties as mentioned in the legend of the figures.
. 
We evaluate the performance of our architecture on value transfer using the Arcade Learning Environment (ALE) platform [\cite{Bellemare}].
\textbf{Atari 2600:}  ALE provides a simulator for Atari 2600 games. This is one of the most commonly used benchmark tasks for deep reinforcement learning algorithms [\cite{mnih2015human}, \cite{mnih2016asynchronous}, \cite{actormimic}, \cite{progressivenets}]. We perform our adaptive transfer learning experiments on the Atari 2600 game Pong.

\subsection{Ability to do Selective Transfer}
In this section, we consider the case when multiple partially favorable source tasks are available such that each of them can assist the learning process for different parts of the state space of the target task. The objective here is to first show the effectiveness of the attention network in learning to {\it focus} only on the source task relevant to the state the agent encounters while trying to complete the target task and then evaluating the full architecture with an additional randomly initialised base network.

\begin{figure}
\centering
\subfloat[The weights given by the attention network. Selective transfer in REINFORCE]{
\label{fig:vis_m2}
\includegraphics[width=0.5\textwidth]{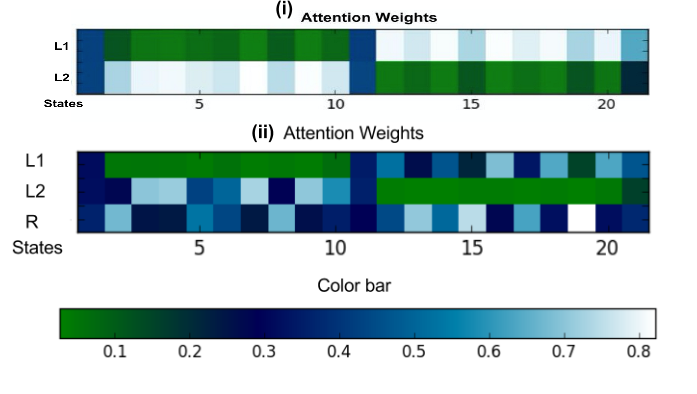} }
\subfloat[Selective transfer in Actor-Critic]{
        \label{fig:sel_transfer}
        \includegraphics[width=0.5\textwidth]{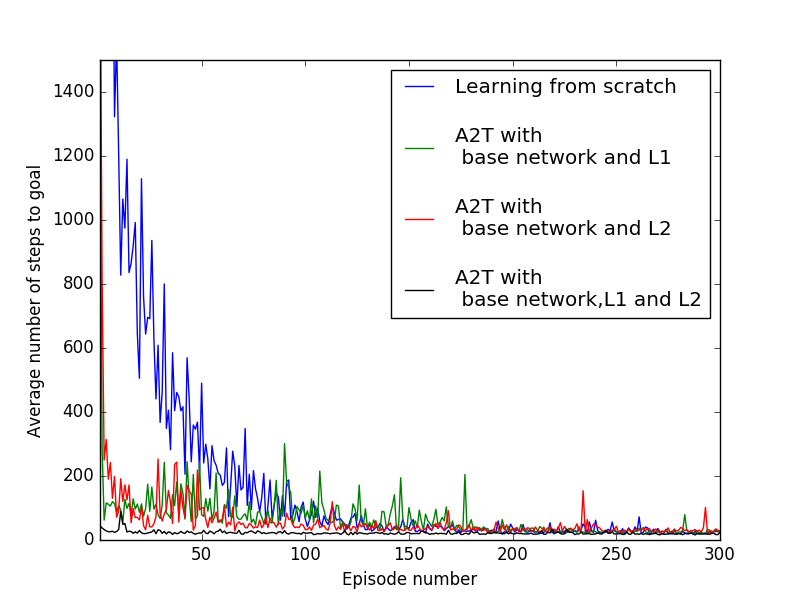} 
        }
\caption{Results of the selective policy transfer experiments}
\end{figure}

This is illustrated for the Policy Transfer setting using the chain world shown in (Fig. \ref{fig:chain_w}). Consider that the target task $LT$ is to start in $A$ or $B$ with uniform probability and reach $C$ in the least number of steps. Now, consider that two learned source tasks, \textit{viz.}, $L1$ and $L2$, are available. $L1$ is the source task where the agent has learned to reach the left end ($A$) starting from the right end ($B$). In contrast, $L2$ is the source task where the agent has learned to reach the right end ($B$) starting from the left end ($A$). Intuitively, it is clear that the target task should benefit from the policies learnt for tasks $L1$ and $L2$. We learn to solve the task $LT$ using REINFORCE given the policies learned for $L1$ and $L2$. Figure \ref{fig:vis_m2} (i) shows the weights given by the attention network to the two source task policies for different parts of the state space at the end of learning. We observe that the attention network has learned to ignore $L1$, and $L2$ for the left, and right half of the state space of the target task, respectively. 
Next, we add base network and evaluate the full architecture on this task. Figure \ref{fig:vis_m2} (ii) shows the weights given by the attention network to the different source policies for different parts of the state space at the end of learning. We observe that the attention network has learned to ignore $L1$, and $L2$ for the left, and right half of the state space of the target task, respectively. As the base network replicates $\pi_T$ over time, it has a high weight throughout the state space of the target task.

We also evaluate our architecture in a relatively more complex puddle world shown in Figure \ref{fig:w2}. In this case, $L1$ is the task of moving from $S1$ to $G1$, and $L2$ is the task of moving from $S2$ to $G1$. In the target task $LT$, the agent has to learn to move to $G1$ starting from either $S1$ or $S2$ chosen with uniform probability. We learn the task $LT$ using Actor-Critic method, where the following are available (i) learned policy for $L1$  (ii) learned policy for $L2$ and (iii) a randomly initialized policy network (the base network). Figure \ref{fig:sel_transfer} shows the performance results. We observe that actor-critic using A2T is able to use the policies learned for $L1$, and $L2$ and performs better than a network learning from scratch without any knowledge of source tasks.

We do a similar evaluation of the attention network, followed by our full architecture for value transfer as well. We create partially useful source tasks through a modification of the Atari 2600 game Pong. We take inspiration from a real world scenario in the sport Tennis, where one could imagine two different right-handed (or left) players with the first being an expert player on the forehand but weak on the backhand, while the second is an expert player on the backhand but weak on the forehand. For someone who is learning to play tennis with the same style (right/left) as the experts, it is easy to follow the forehand expert player whenever he receives a ball on the forehand and follow the backhand expert whenever he receives a ball on the backhand.
\begin{figure}
\centering
\includegraphics[width=0.9\textwidth]{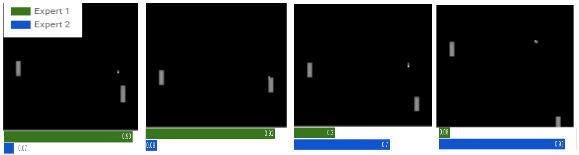}
 \caption{Visualisation of the attention weights in the Selective Transfer with Attention Network experiment: Green and Blue bars signify the attention probabilities for Expert-1 ($L1$) and Expert-2 ($L2$) respectively. We see that in the first two snapshots, the ball is in the lower quadrant and as expected, the attention is high on Expert-1, while in the third and fourth snapshots, as the ball bounces back into the upper quadrant, the attention increases on Expert-2.}
\label{fig:pong_att_vis}
\end{figure}

We try to simulate this scenario in Pong. The trick is to blur the part of the screen where we want to force the agent to be weak at returning the ball.
The blurring we use is to just black out all pixels in the specific region required. To make sure the blurring doesn't contrast with the background, we modify Pong to be played with a black background (pixel value $0$) instead of the existing gray (pixel value $87$). We construct two partially helpful source task experts $L1$ and $L2$. $L1$ is constructed by training a DQN on Pong with the upper quadrant (the agent's side) blurred, while $L2$ is constructed by training a DQN with the lower quadrant (the agent's side) blurred. This essentially results in the ball being invisible when it is in the upper quadrant for $L1$ and lower quadrant for $L2$.  We therefore expect $L1$ to be useful in guiding to return balls on the lower quadrant, and $L2$ for the upper quadrant. The goal of the attention network is to learn suitable filters and parameters so that it will focus on the correct source task for a specific situation in the game.
The source task experts $L1$ and $L2$ scored an average
of {\textbf{9.2}} and {\textbf{8}} respectively on Pong game play with black background. With an attention network to suitably weigh the value functions of $L1$ and $L2$, an average performance of {\textbf{17.2}} was recorded just after a single epoch (250,000 frames) of training. (The score in Pong is in the range of $[-21,21]$). This clearly shows that the attention mechanism has learned to take advantage of the experts adaptively. Fig. \ref{fig:pong_att_vis} shows a visualisation of the attention weights for the same.

\begin{wrapfigure}{r}{0.48\textwidth}
    \centering
    \includegraphics[width=0.52\textwidth]{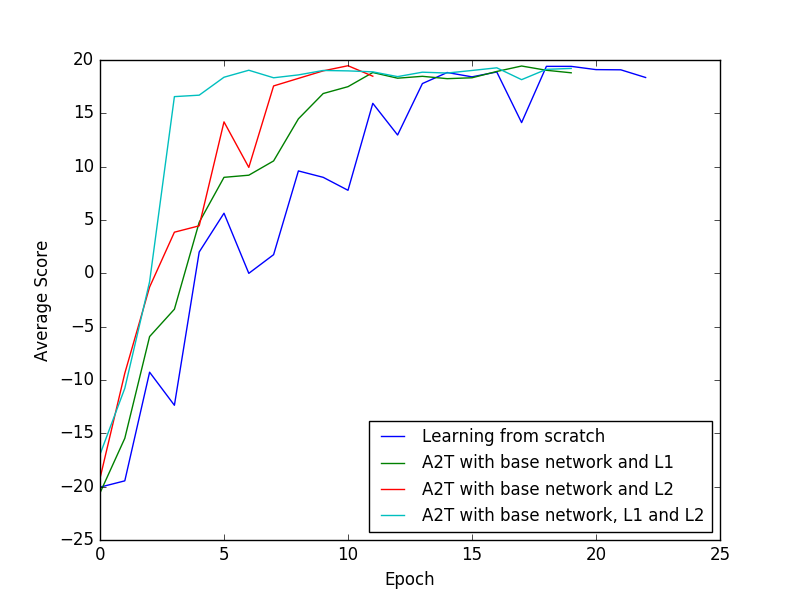}
    \caption{Selective Value Transfer.}
    \vspace{-10pt}
    \label{fig:2selective_blurring}
\end{wrapfigure}

We then evaluate our full architecture (A2T) in this setting, i.e with an addition of DQN learning from scratch (base network) to the above setting. The architecture can take advantage of the knowledge of the source task experts selectively early on during the training while using the expertise of the base network wherever required, to perform well on the target task. Figure \ref{fig:2selective_blurring} summarizes the results, where it is clear that learning with both the partially useful experts is better than learning with only one of them which in turn is better than learning from scratch without any additional knowledge.

\subsection{Ability to Avoid Negative Transfer and Ability to Transfer from Favorable Task}
We first consider the case when only one learned source task is available such that its solution $K_{1}$ (policy or value) can hamper the learning process of the new target task. We refer to such a source task as an unfavorable source task. In such a scenario, the attention network shown in Figure \ref{fig:2m2} 
should learn to assign a very low weight (ignore) to $K_{1}$ . 
We also consider a modification of this setting by adding another source task whose solution $K_2$ is favorable to the target task. In such a scenario, the attention network should learn to assign high weight (attend) to $K_2$ while ignoring $K_1$.

\begin{wrapfigure}{r}{0.48\textwidth}
    \centering
    \includegraphics[width=0.52\textwidth]{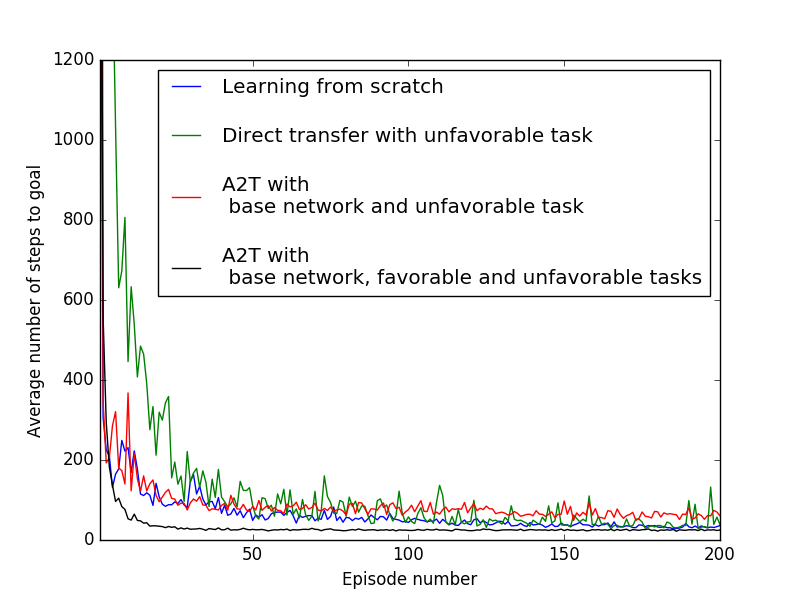}
    \caption{Avoiding negative transfer and transferring policy from a favorable task(lower the better).}
    \label{fig:4neg_transfer_fav_task}
\end{wrapfigure}

We now define an experiment using the puddle world from Figure \ref{fig:w1} for policy transfer. The target task in our experiment is to maximize the return in reaching the goal state $G1$ starting from any one of the states ${S1, S2, S3, S4}$. We artificially construct an unfavorable source task by first learning to solve the above task and then negating the weights of the topmost layer of the actor network. We then add a favorable task to the above setting. 
We artificially construct a favorable source task simply by learning to solve the target task and using the learned actor network. Figure \ref{fig:4neg_transfer_fav_task} shows the results. The target task for the value transfer experiment is to reach expert level performance on Pong. We construct two kinds of unfavorable source tasks for this experiment. \textbf{Inverse-Pong}: A DQN on Pong trained with negated reward functions, that is with $R'(s,a) = - R(s,a)$ where $R(s,a)$ is the reward provided by the ALE emulator for choosing action $a$ at state $s$. \textbf{ Freeway}: An expert DQN on another Atari 2600 game, Freeway, which has the same range of optimal value functions and same action space as Pong. We empirically verified that the Freeway expert DQN leads to negative transfer when directly initialized and fine-tuned on Pong which makes this a good proxy for a negative source task expert even though the target task Pong has a different state space.

\begin{figure}
\centering
\subfloat[Avoiding negative transfer(Pong) and transferring from a favorable task]{
        \label{fig:2selective_pong}
        \includegraphics[width=0.47\textwidth]{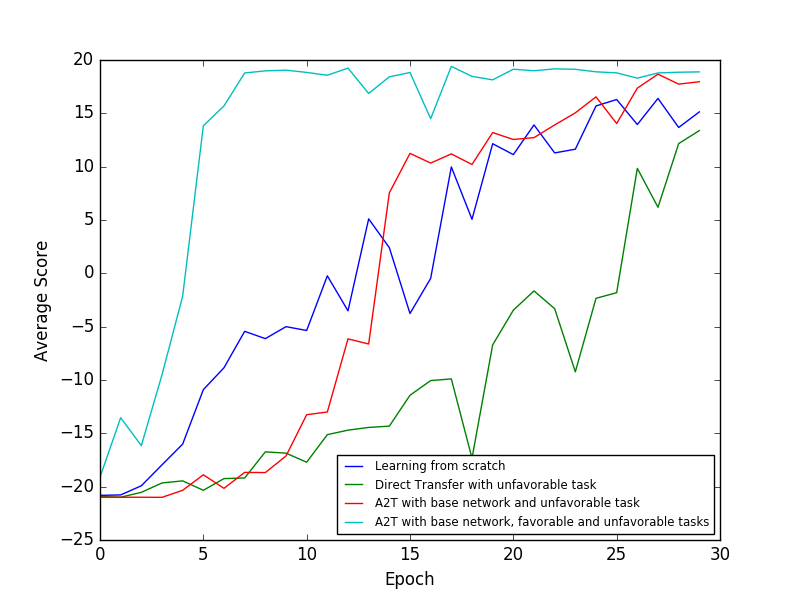}}
\subfloat[Avoiding negative transfer(Freeway) and transferring from a favorable task]{
        \label{fig:2selective_freeway}
        \includegraphics[width=0.47\textwidth]{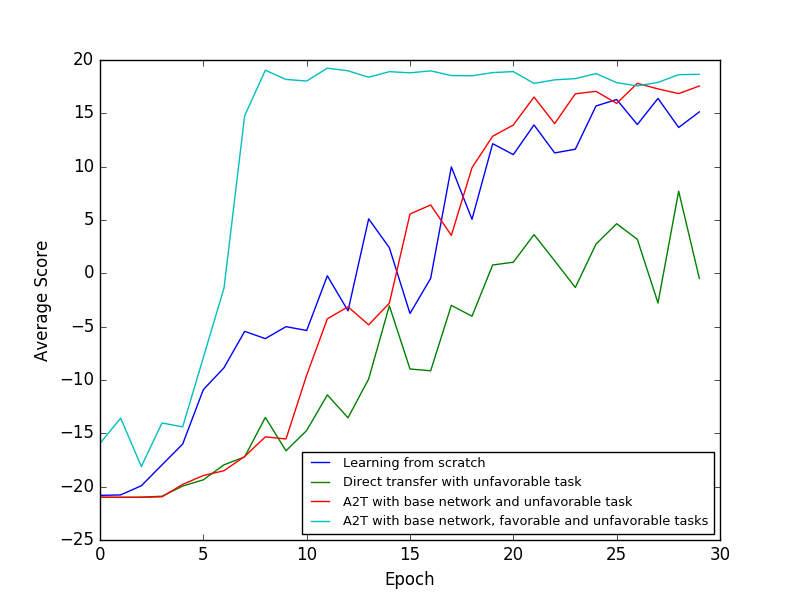} }
\caption{Avoiding negative transfer and transferring value from a favorable task(higher the better). Specific training and architecture details are mentioned in APPENDIX. The plots are averaged over two runs with different random seeds.}
\end{figure}
We artificially construct a favorable source task by learning a DQN to achieve expertise on the target task (Pong) and use the learned network. Figure \ref{fig:2selective_pong} compares the performance of the various scenarios when the unfavorable source task is Inverse-Pong, while Figure  \ref{fig:2selective_freeway} offers a similar comparison with the negative expert being Freeway. 



From all the above results, we can clearly see that A2T does not get hampered by the unfavorable source task by learning to ignore the same and performs competitively with just a randomly initialized learning on the target task without any expert available. Secondly, in the presence of an additional source task that is favorable, A2T learns to transfer useful knowledge from the same while ignoring the unfavorable task, thereby reaching expertise on the target task much faster than the other scenarios.

\subsection{Visualization: Evolution of Attention Weights with one positive and one negative expert}

\begin{wrapfigure}{r}{0.48\textwidth}
    \centering
    \includegraphics[width=0.52\textwidth]{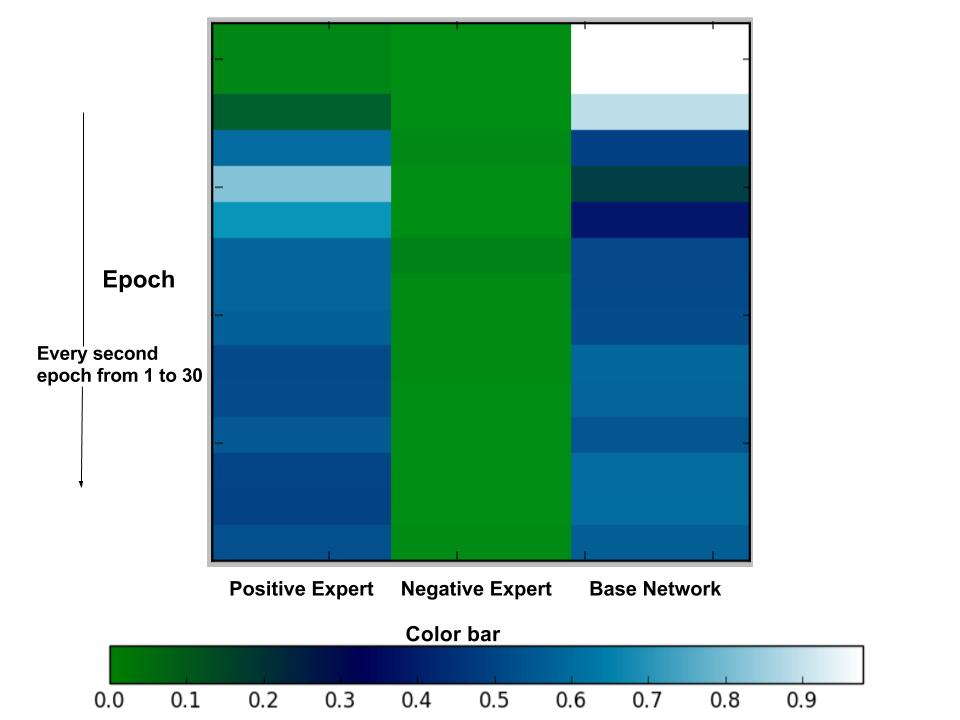}
    \caption{Evolution of attention weights with one positive and one negative expert.}
    \label{fig:4att_viz}
\end{wrapfigure}

We present the evolution of attention weights for the experiment described in Section 4.2 where we focus on the efficacy of the A2T framework in providing an agent the ability to {\it avoid negative transfer} and {\it transfer from a favorable source task (perfect expert)}. Figure \ref{fig:4att_viz} depicts the evolution of the attention weights (normalised in the range of $[0,1]$) during the training of the A2T framework. The corresponding experiment is the case where the target task is to solve Pong, while there are two source task experts, one being a perfect Pong playing trained DQN (to serve as positive expert), and the other being the {\it Inverse-Pong} DQN trained with negated reward functions (to serve as negative expert). Additionally, there's also the base network that learns from scratch using the experience gathered by the attentively combined behavioral policy from the expert networks, the base network and itself.

We train the framework for 30 epochs, and the plot illustrates the attention weights every second epoch. We clearly see from figure \ref{fig:4att_viz} that there is no weird co-adaptation that happens in the training, and the attention on the negative expert is uniformly low throughout. Initially, the framework needs to collect some level of experience to figure out that the positive expert is optimal (or close to optimal). Till then, the attention is mostly on the base network, which is learning from scratch. The attention then shifts to the positive expert which in turn provides more rewarding episodes and transition tuples to learn from. Finally, the attention drifts slowly to the base network from the positive expert again, after which the attention is roughly random in choosing between the execution of positive expert and the base network. This is because the base network has acquired sufficient expertise as the positive expert which happens to be optimal for the target task. This visualization clearly shows that A2T is a powerful framework in ignoring a negative expert throughout and using a positive expert appropriately to learn quickly from the experience gathered and acquire sufficient expertise on the target task.

\subsection{When a perfect expert is not available among the source tasks}
\begin{wrapfigure}{r}{0.48\textwidth}
    \centering
    \includegraphics[width=0.52\textwidth]{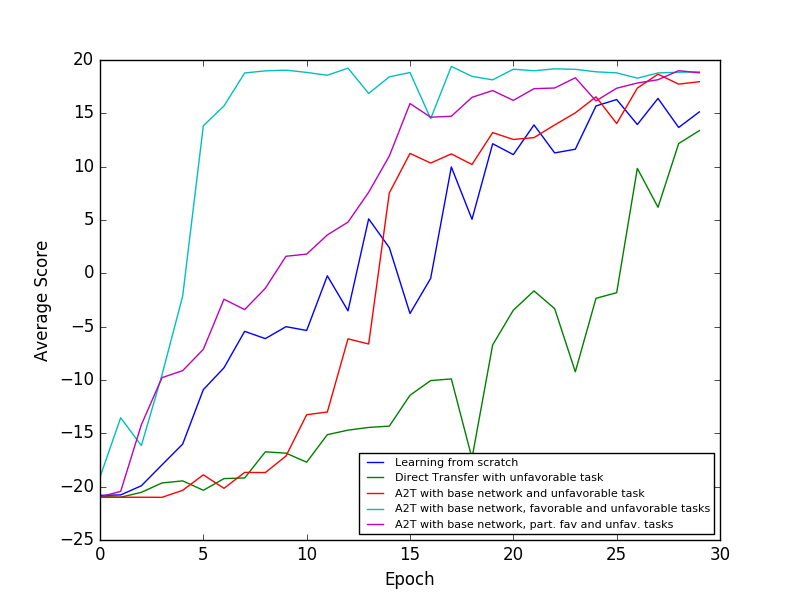}
    \caption{Partial Positive Expert Experiment}
    \label{fig:4partial_pos}
\end{wrapfigure}
    
In our experiments in the previous subsection dealing with prevention of negative transfer and using a favorable source task, we consider the positive expert as a perfect (close to optimal) expert on the same task we treat as the target task. This raises the question of relying on the presence of a perfect expert as a positive expert. If we have such a situation, the obvious solution is to execute each of the experts on the target task and vote for them with probabilities proportional to the average performance of each. 

The A2T framework is however generic and not intended to just do {\it source task selection}. We illustrate this with an additional baseline experiment, where the positive source task is an {\it imperfect expert on the target task}. In such a case, just having a weighted average voting among the available source task networks based on their individual average rewards is upper bounded by the performance of the best available positive expert, which happens to be an imperfect expert on the target task. Rather, the base network has to acquire new skills not present in the source task networks. We choose a partially trained network on Pong, that scores an average of {\textbf{8}} (max: 21). The graph in figure \ref{fig:4partial_pos} clearly shows that the A2T framework with a partial Pong expert and a negative expert performs better than i) learning from scratch, ii) A2T with only one negative expert, and performs worse than A2T with one {\it perfect} positive expert and one negative expert. This is expected because a partial expert cannot provide as much of expert knowledge as a perfect expert, but still provides some useful knowledge in speeding the process of solving the target task. An important conclusion from this experiment is that the A2T framework is capable of discovering new skills not available among any of the experts when such skills are required for optimally solving the target task. To maintain consistency, we perform the same number of runs for averaging scores and experimented with both learning rates and pick the better performing one (0.00025).


\section{Conclusion and Future work}
In this paper we present a very general deep neural network architecture, A2T, for transfer learning that avoids negative transfer while enabling selective transfer from multiple source tasks in the same domain. We show simple ways of using A2T for policy transfer and value transfer. We empirically evaluate its performance with different algorithms, using simulated worlds and games, and show that it indeed achieves its stated goals. Apart from transferring task solutions, A2T can also be used for transferring other useful knowledge such as the model of the world.

While in this work we focused on transfer between tasks that share the same state and action spaces and are in the same domain, the use of deep networks opens up the possibility of going beyond this setting. For example, a deep neural network can be used to learn common representations [\cite{actormimic}] for multiple tasks thereby enabling transfer between related tasks that could possibly have different state-action spaces. A hierarchical attention over the lower level filters across source task networks while learning the filters for the target task network is another natural extension to transfer across tasks with different state-action spaces. The setup from Progressive Neural Networks [\cite{progressivenets}] could be borrowed for the filter transfer, while the A2T setup can be retained for the policy/value transfer. Exploring this setting for continuous control tasks so as to transfer from modular controllers as well avoid negative transfer is also a potential direction for future research.

The nature of tasks considered in our experiments is naturally connected to Hierarchical Reinforcement Learning and Continual Learning. For instance, the blurring experiments inspired from Tennis based on experts for specific skills like Forehand and Backhand could be considered as learning from sub-goals (program modules) like Forehand and Backhand to solve a more complex and broader task like Tennis by invoking the relevant sub-goals (program modules). This structure could be very useful to build a household robot for general purpose navigation and manipulation whereby specific skills such as manipulation of different objects, navigating across different source-destination points, etc could be invoked when necessary. The attention network in the A2T framework is essentially a {\it soft meta-controller} and hence presents itself as a powerful {\it differentiable} tool for Continual and Meta Learning. Meta-Controllers have typically been been designed with discrete decision structure over high level subgoals. This paper presents an alternate differentiable meta-controller with a soft-attention scheme. We believe this aspect can be exploited for differentiable meta-learning architectures for hierarchical reinforcement learning. 
Over all, we believe that A2T is a novel way to approach different problems like Transfer Learning, Meta-Learning and Hierarchical Reinforcement Learning and further refinements on top of this design can be a good direction to explore.

\section*{Acknowledgements}
Thanks to the anonymous reviewers of ICLR 2017 who have provided thoughtful remarks and helped us revise the paper. We would also like to thank Sherjil Ozair, John Schulman, Yoshua Bengio, Sarath Chandar, Caglar Gulchere and Charu Chauhan for useful feedback about the work. 
\newpage

\bibliography{iclr2017_conference}
\bibliographystyle{iclr2017_conference}

\newpage

\section*{APPENDIX A: Details of the Network Architecture in Value Transfer Experiments}
For the source task expert DQNs, we use the same architecture as [\cite{mnih2015human}] where the input is $84\times 84\times 4$ with $32$ convolution filters, dimensions $8\times 8$, stride $4\times 4$ followed by $64$ convolution filters with dimensions $4\times 4$ and stride $2\times 2$, again followed by $64$ convolution filters of size $3\times 3$ and stride $1\times 1$. This is then followed by a fully connected layer of $512$ units and finally by a fully connected output layer with as many units as the number of actions in Pong (Freeway) which is $3$. We use ReLU nonlinearity in all the hidden layers.

With respect to the A2T framework architecture, we have experimented with two possible architectures:

\begin{itemize}
    \item The base and attention networks following the NIPS architecture of \cite{mnih2013playing} except that the output layer is softmax for the attention network. 
    \item The base and attention networks following the Nature architecture of \cite{mnih2015human} with a softmax output layer for the attention network.
\end{itemize}

Specifically, the NIPS architecture of \cite{mnih2013playing} takes in a batch of $84 \times 84 \times 4$ inputs, followed by $16$ convolution filters of dimensions $8 \times 8$ with stride $4 \times 4$, $32$ convolution filters with dimensions $4\times 4$ and stride $2\times 2$, a fully connected hidden layer of $256$ units, followed by the output layer. For the Selective Transfer with Blurring experiments described in Section 4.1, we use the second option above. For the other experiments in Section 4.2 and the additional experiments in Appendix, we use the first option. The attention network has $N+1$ outputs where $N$ is the number of source tasks.

\section*{APPENDIX B: Training Details }
\subsection*{Training Algorithm}

For all our experiments in Value Transfer, we used RMSProp as in [\cite{mnih2015human}] for updating gradient. For Policy Transfer, since the tasks were simple, stochastic gradient descent was sufficient to provide stable updates. We also use reward clipping, target networks and experience replay for our value transfer experiments in exactly the same way (all hyper parameters retained) as [\cite{mnih2015human}]. A training epoch is 250,000 frames and for each training epoch, we evaluate the networks with a testing epoch that lasts 125,000 frames. We report the average score over the completed episodes for each testing epoch. The average scores obtained this way are averaged over 2 runs with different random seeds. In the testing epochs, we use $\epsilon=0.05$ in the $\epsilon$-greedy policy.

\subsection*{Learning Rate}

In all our experiments, we trained the architecture using the learning rates, $0.0025$ and $0.0005$. In general, the lower learning rate provided more stable (less variance) training curves. While comparing across algorithms, we picked the best performing learning rate out of the two ($0.0025$ and $0.0005$) for each training curve.

\section*{APPENDIX C: Blurring Experiments on Pong}

The experts are trained with blurring (hiding the ball) and black background as illustrated in APPENDIX A. Therefore, to compare the learning with that of a random network without any additional knowledge, we ran the baseline DQN on Pong with a black background too. Having a black background provides a rich contrast between the white ball and the black background, thereby making training easier and faster, which is why the performance curves in that setting are different to the other two settings reported for Inverse Pong and Freeway Negative transfer experiments where no blacking is done and Pong is played with a gray background. The blurring mechanism in Pong is illustrated in APPENDIX E.

\begin{figure*}[t]
\centering
\section*{APPENDIX E: Blurring Mechanism in Pong - Details}
\subfloat[Ball in upper quad]{
        \label{fig:2pong_up}
        \includegraphics[width=0.2\textwidth]{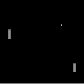}}
\subfloat[Blurred upper quad]{
        \label{fig:2pong_up_blacked}
        \includegraphics[width=0.2\textwidth]{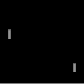} }
\subfloat[Ball in lower quad]{
        \label{fig:2pong_down}
        \includegraphics[width=0.2\textwidth]{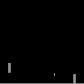}}
\subfloat[Blurred lower quad]{
        \label{fig:2pong_down_blacked}
        \includegraphics[width=0.2\textwidth]{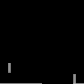}}\\
\caption{The figures above explain the blurring mechanism for selective transfer experiments on Pong. The background of the screen is made black. Let $X$ ($84 \times 84$) denote an array containing the pixels of the screen. The paddle controlled by the agent is the one on the right. We focus on the two quadrants $X1 = X[:42, 42:]$ and $X2 = X[42:,42:]$ of the Pong screen relevant to the agent controlled paddle. To simulate an expert that is weak at returning balls in the upper quadrant, the portion of $X1$ till the horizontal location of agent-paddle, ie $X1[:,:31]$ is blacked out, while similarly, for simulating weakness in the bottom quadrant, we blur the portion of $X2$ till the agent-paddle's horizontal location, ie $X2[:,:31]=0$. Figures \ref{fig:2pong_up} and \ref{fig:2pong_up_blacked} illustrate the scenarios of blurring the upper quadrant before and after blurring; and similarly do \ref{fig:2pong_down} and \ref{fig:2pong_down_blacked} for blurring the lower quadrant. Effectively, blurring this way with a black screen is equivalent to hiding the ball (white pixel) in the appropriate quadrant where weakness is to be simulated. Hence, Figures \ref{fig:2pong_up_blacked} and \ref{fig:2pong_down_blacked} are the mechanisms used while training a DQN on Pong to hide the ball at the respective quadrants, so to create the partially useful experts which are analogous to forehand-backhand experts in Tennis. $X[:a,:b]$ indicates the subarray of $X$ with all rows upto row index $a$ and all columns upto column index $b$.} 
\label{fig:pong_blurring}
\end{figure*}

\section*{APPENDIX D: Blurring experiments on Breakout}

Similar to our Blurring experiment on Pong, we additionally ran another experiment on the Atari 2600 game, Breakout, to validate the efficiency of our attention mechanism. We consider a setup with two experts $L1$ and $L2$ along with our attention network. The experts $L1$ and $L2$ were trained by blurring the lower left and right quadrants of the breakout screen respectively. We don't have to make the background black like in the case of Pong because the background is already black in Breakout and direct blurring is sufficient to hiding the ball in the respective regions without any contrasts introduced. We blur only the lower part so as to make it easy for the agent to at least anticipate the ball based on the movement at the top. We empirically observed that blurring the top half (as well) makes it hard to learn any meaningful partially useful experts $L1$ and $L2$. 

The goal of this experiment is to show that the attention network can learn suitable filters so as to dynamically adapt and learn to select the expert appropriate to the situation (game screen) in the task. The expert $L1$ which was blurred on the left bottom half is bound to weak at returning balls on that region while $L2$ is expected to be weak on the right. This is in the same vein as the forehand-backhand example in Tennis and its synthetic simulation for Pong by blurring the upper and lower quadrants. During game play, the attention mechanism is expected to ignore $L2$ when the ball is on the bottom right half (while focusing on $L1$) and similarly ignore $L2$ (while focusing on $L1$) when the ball is on the left bottom half. We learn experts $L1$ and $L2$ which score {\textbf{42.2}} and {\textbf{39.8}} respectively. Using the attention mechanism to select the correct expert, we were able to achieve a score of {\textbf{94.5}} after training for $5$ epochs. Each training epoch corresponds to $250,000$ decision steps, while the scores are averaged over completed episodes run for $125,000$ decision steps. This shows that the attention mechanism learns to select the suitable expert. Though the performance is limited by the weaknesses of the respective experts, our goal is to show that the attention paradigm is able to take advantage of both experts appropriately. This is evident from the scores achieved by standalone experts and the attention mechanism. Additionally, we also present a visualization of the attention mechanism weights assigned to the experts $L1$ and $L2$ during game play in APPENDIX G. The weights assigned are in agreement with what we expect in terms of selective attention. The blurring mechanism is visually illustrated in APPENDIX F.

\begin{figure*}
\centering
\section*{APPENDIX F: Blurring Mechanism in Breakout - Details}

\subfloat[Ball in lower-left quad]{
        \label{fig:2breakout_up}
        \includegraphics[width=0.25\textwidth]{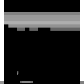}}
\subfloat[Blurred lower-left quad]{
        \label{fig:2breakout_up_blacked}
        \includegraphics[width=0.25\textwidth]{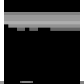} }
\subfloat[Ball in  lower-right quad]{
        \label{fig:2breakout_down}
        \includegraphics[width=0.25\textwidth]{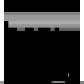}}
\subfloat[Blurred lower-right quad]{
        \label{fig:2breakout_down_blacked}
        \includegraphics[width=0.25\textwidth]{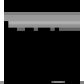}}
\caption{The figures above explain the blurring mechanism used for selective transfer experiments on Breakout. The background of the screen is already black. Let $X$ ($84 \times 84$) denote an array containing the pixels of the screen. We focus on the two quadrants $X1 = X[31:81,4:42]$ and $X2=X[31:81,42:80]$. We perform blurring in each case by ensuring $X1=0$ and $X2=0$ for all pixels within them for training $L1$ and $L2$ respectively. Effectively, this is equivalent to hiding the ball in the appropriate quadrants. Blurring $X1$ simulates weakness in the lower left quadrant, while blurring $X2$ simulates weakness in the lower right quadrant. We don't blur all the way down upto the last row to ensure the paddle controlled by the agent is visible on the screen. We also don't black the rectangular border with a width of $4$ pixels surrounding the screen.  Figures \ref{fig:2breakout_up} and \ref{fig:2breakout_up_blacked} illustrate the scenarios of blurring the lower left quadrant before and after blurring; and similarly do \ref{fig:2breakout_down} and \ref{fig:2breakout_down_blacked} for blurring the lower right quadrant.}
\label{fig:breakout_blurring}
\end{figure*}

\begin{figure*}
\section*{APPENDIX G: Blurring Attention Visualization on Breakout}
\centering
\includegraphics[width=0.95\textwidth]{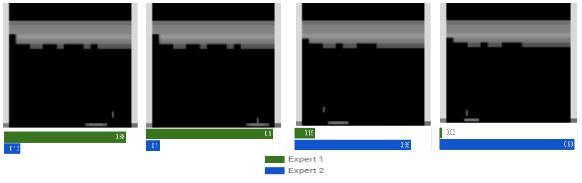}
 \caption{Visualisation of the attention weights in the Selective Transfer with Attention for Breakout: Green and Blue bars signify the attention probabilities for Expert-1 ($L1$) and Expert-2 ($L2$) respectively on a scale of $[0,1]$. We see that in the first two snapshots, the ball is in the lower right quadrant and as expected, the attention is high on Expert-1, while in the third and fourth snapshots, the ball is in the lower right quadrant and hence the attention is high on Expert-2.}
\label{fig:breakout_att_vis}
\end{figure*}

\begin{figure*}
    \section*{APPENDIX J: Case study of target task performance limited by data availability}
    \centering
    
    \subfloat[Comparison of Sparse Pong to Normal Pong]{
        \label{fig:sparse_pong}
        \includegraphics[width=0.45\textwidth]{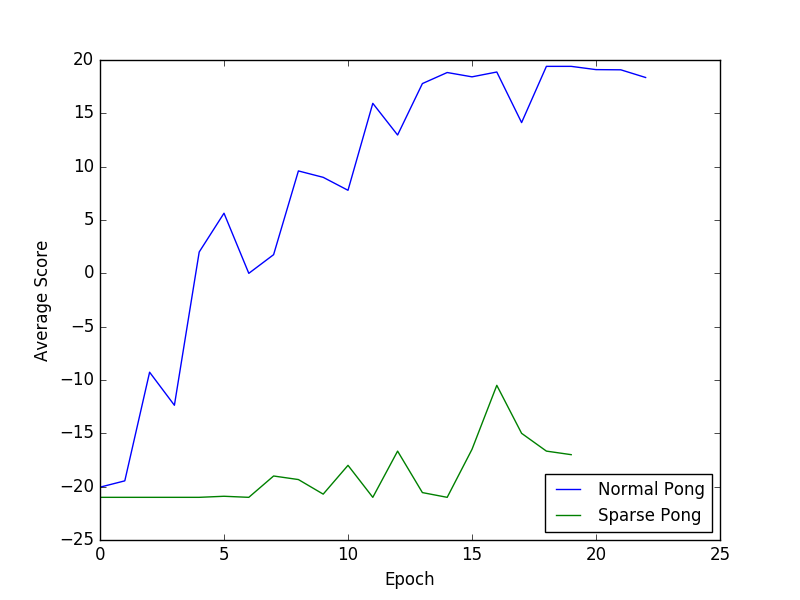}}
\subfloat[A2T with a positive and negative expert]{
        \label{fig:sparse_pong_a2t}
        \includegraphics[width=0.45\textwidth]{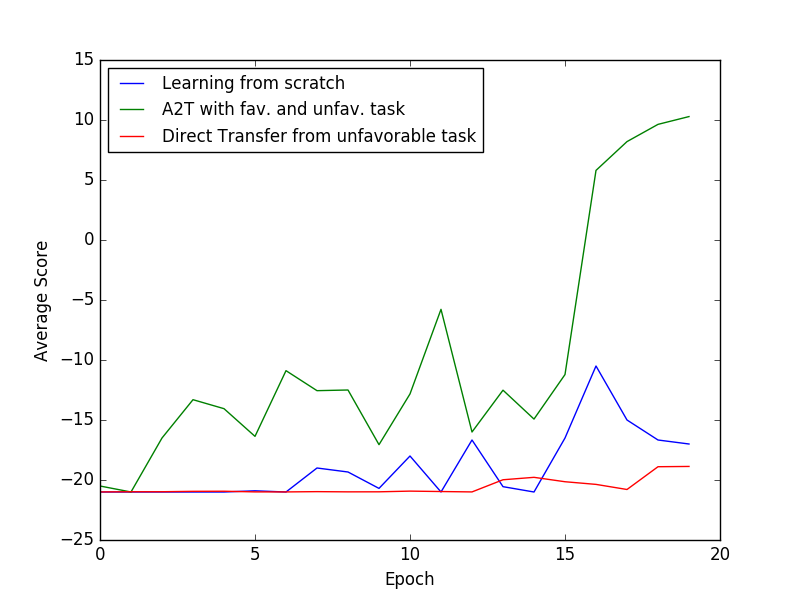} }
    \caption{This experiment is a case study on a target task where the performance is limited by data availability. So far, we focused on experiments where the target task is to solve Pong (normal or black background) for Value Transfer, and Puddle Worlds for Policy Transfer. In both these cases, a randomly initialized value (or policy) network learning without the aid of any expert network is able to solve the target task within a reasonable number of epochs (or iterations). We want to illustrate a case where solving the target task in reasonable time is hard and the presence of a favorable source task significantly impacts the speed of learning. To do so, we consider a variant of Pong as our target task. In this variant, only a small probability $\rho$ of transition tuples $(s,a,r,s')$ with {\it non-zero reward} $r$ are added to the Replay Memory (and used for learning through random batch sampling). This way, the performance on the target task is limited by the availability of rewarding (positive or negative) transitions in the replay memory. This synthetically makes the target task of Pong a sparse reward problem because the replay memory is largely filled with transition tuples that have zero reward. We do not use any prioritized sampling so as to make sure the sparsity has a negative effect on learning to solve the target task. We use a version of Pong with black background (as used in Section 4.1 for the Blurring experiments) for faster experimentation. $\rho = 0.1$ was used for the plots illustrated above. Figure \ref{fig:sparse_pong} clearly shows the difference between a normal Pong task without any synthetic sparsity and the new variant we introduce. The learning is much slower and is clearly limited by data availability even after 20 epochs (20 million frames) due to reward sparsity. Figure \ref{fig:sparse_pong_a2t} describes a comparison between the A2T setting with one positive expert which expertly solves the target task and one negative expert, learning from scratch, and direct fine-tuning on a negative expert. We clearly see the effect of having the positive expert in one of the source tasks speeding up the learning process significantly when compared to learning from scratch, and also see that fine-tuning on top of a negative expert severely limits learning even after 20 epochs of training. We also see that the A2T framework is powerful to work in sparse reward settings and avoids negative transfer even in such cases, while also clearly learning to benefit from the presence of a target task expert among the source task networks. Importantly, this experiment demonstrates that transfer learning has a significant effect on tasks which may be hard (infeasible to solve within a reasonable training time) without any expert available. Further, A2T is also beneficial for such (sparse reward) situations when accessing the weights of an expert network is not possible, and only outputs of the expert (policy or value-function) can be used. Such synthetic sparse variants of existing tasks is a good way to explore future directions in the intersection of Inverse Reinforcement Learning and Reward-Based Learning, with A2T providing a viable framework for off-policy and on-policy learning.}
    \label{fig:sparse_pong_experiments}
\end{figure*}

\end{document}